\documentclass[11pt,a4paper]{article}
\usepackage[hyperref]{style/acl2021}
\usepackage{times}
\usepackage{latexsym}

\usepackage{microtype}

\aclfinalcopy % Uncomment this line for the final submission
\def\aclpaperid{361} %  Enter the acl Paper ID here

\title{DoT: An efficient Double Transformer for NLP tasks with tables}

\author{
Syrine Krichene$^1$,
Thomas M{\"u}ller$^2$\thanks{\,\,\,Work done at Google Research.},
Julian Martin Eisenschlos$^1$\\
Google Research, Z{\"u}rich$^1$ \\
\texttt{\{syrinekrichene,eisenjulian\}@google.com}\\
Symanto Research, Valencia, Spain$^2$ \\
\texttt{thomas.mueller@symanto.com}
}

\usepackage{amsthm}
\usepackage{framed}
\usepackage{mdwlist}
\usepackage{colortbl}
\usepackage{xcolor}
\usepackage{nicefrac}
\usepackage{booktabs}
\usepackage{amsfonts}
\usepackage[T1]{fontenc}
\usepackage{bold-extra}
\usepackage{amsmath}
\usepackage{amssymb}
\usepackage{bm}
\usepackage{graphicx}
\usepackage{mathtools}
\usepackage{multirow}
\usepackage{multicol}
\usepackage[normalem]{ulem}
\usepackage{lipsum}
\usepackage{float}
\usepackage{ifthen}
\usepackage[ruled,vlined]{algorithm2e}
\usepackage{subcaption}
\usepackage{tabularx}
\usepackage[utf8]{inputenc}
\usepackage[english]{babel}

\newtheorem{theorem}{Theorem}[section]

\newtheorem*{remark}{Remark}

\usepackage{tikz}
\usepackage{xspace}
\usepackage{xr}
\usetikzlibrary{shapes,arrows}

\newcommand{\gem}[1]{\mbox{\textsc{gem}}}
\newcommand{\abr}[1]{\textsc{#1}}

\newcommand{\hidetext}[1]{}
\newcommand{\ignore}[1]{}

\newif\ifinfotabs\infotabsfalse

\newif\ifsubscripterror\subscripterrortrue

\ifinfotabs
\newcommand{\infotabstext}[1]{#1}
\else
\newcommand{\infotabstext}[1]{}
\fi

\ifsubscripterror

\else

\fi

\newcommand{\smallurl}[1]{ \begin{tiny}\url{#1}\end{tiny}}

\definecolor{lightblue}{HTML}{3cc7ea}
\definecolor{CUgold}{HTML}{CFB87C}
\definecolor{grey}{rgb}{0.95,0.95,0.95}
\definecolor{ceil}{rgb}{0.57, 0.63, 0.81}
\definecolor{UMDred}{HTML}{ed1c24}
\definecolor{UMDyellow}{HTML}{ffc20e}
\definecolor{darkgreen}{HTML}{008f00}

\newcommand{\tabfact}{\textsc{TabFact}\xspace}

\newcommand\sota{state-of-the-art\xspace}
\newcommand{\tapas}{\textsc{TAPAS}\xspace}
\newcommand{\wtq}{\textsc{WIKITQ}\xspace}
\newcommand{\wikisql}{\textsc{WIKISQL}\xspace}
\newcommand{\bert}{\textsc{BERT}\xspace}

\theoremstyle{definition}

\makeatletter
\newcommand*{\addFileDependency}[1]{% argument=file name and extension
  \typeout{(#1)}
  \@addtofilelist{#1}
  \IfFileExists{#1}{}{\typeout{No file #1.}}
}
\makeatother

\begin{document}
\maketitle

\begin{abstract}
    Transformer-based approaches have been successfully used to obtain state-of-the-art accuracy on natural language processing (NLP) tasks with  semi-structured tables. These model architectures are typically deep, resulting in slow training and inference, especially for long inputs. To improve efficiency while maintaining a high accuracy, we propose a new architecture, $DoT$, a double transformer model, that decomposes the problem into two sub-tasks: A shallow pruning transformer that selects the top-$K$ tokens, followed by a deep task-specific transformer that takes as input those K tokens. Additionally, we modify the task-specific attention to incorporate the pruning scores. The two transformers are jointly trained by optimizing the task-specific loss. We run experiments on three benchmarks, including entailment and question-answering. We show that for a small drop of accuracy, $DoT$ improves training and inference time by at least 50\%. We also show that the pruning transformer effectively selects relevant tokens enabling the end-to-end model to maintain similar accuracy as slower baseline models. Finally, we analyse the pruning and give some insight into its impact on the task model.
\end{abstract}

\section{Introduction}

Recently, transfer learning with large-scale pre-trained language models has been successfully used to solve many NLP tasks \cite{devlin-etal-2019-bert,radford2019rewon,liu2019roberta}. In particular, transformer models have been used to solve tasks that include semi-structured table knowledge, such as table question answering~\cite{herzig-etal-2020-tapas} and entailment~\cite{2019TabFactA,eisenschlos2020understanding} -- a binary classification task to support or refute a sentence based on the table's content.

While transformer models lead to significant improvements in accuracy, they suffer from high computation and memory cost, especially for large inputs.  The total computational complexity per layer for self-attention is $O(n^2 d)$ \cite{vaswani2017attention}, where $n$ is the input sequence length, and $d$ is the embedding dimension. Using longer sequence lengths translates into increased training and inference time. 

Improving the computational efficiency of transformer models has recently become an active research topic. To the best of our knowledge, the only technique that was applied to \abr{nlp} tasks with semi-structured tables is heuristic pruning.
\citet{eisenschlos2020understanding} show on the \tabfact data set~\cite{2019TabFactA} that using heuristic pruning accelerates the training time while achieving a similar accuracy. 
This raises the question of whether a better pruning strategy can be learned.

\begin{figure}
    \centering
    \scalebox{1.07}{
    \hspace{-2.3ex}
    \includegraphics[width=1.0\linewidth]{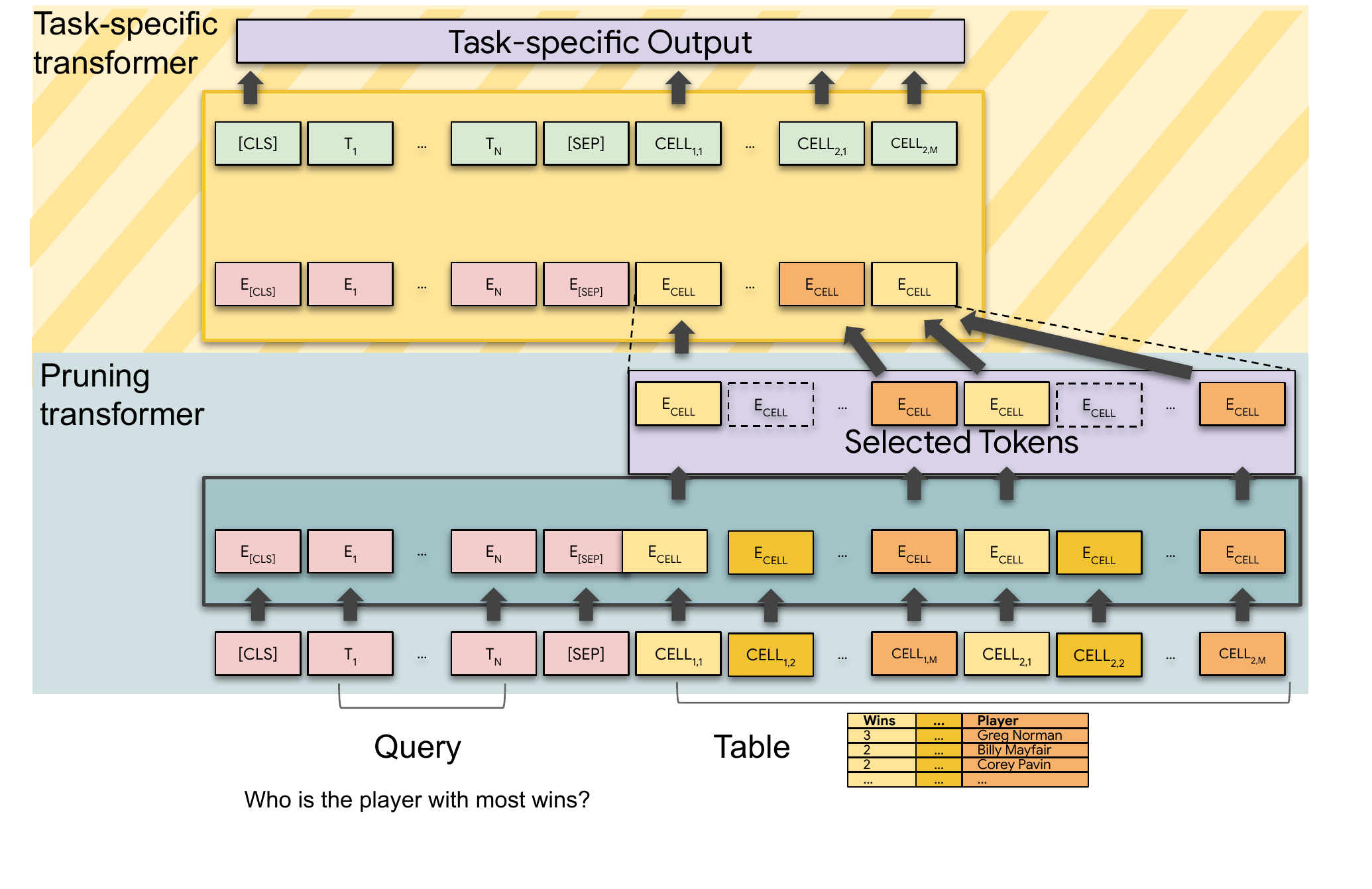}}
    \vspace{-7ex}
    \caption{Pruning with a double transformer $DoT$. The pruning model selects the $k$ most relevant tokens and passes them to the task model. The pruning model is small, allowing the use of long input sequences.}
    \label{fig:pruning_model}
    % \vspace{-1.0ex}
\end{figure}

We propose to use $DoT$, a double transformer model (Figure~\ref{fig:pruning_model}): A first transformer -- which we call \emph{pruning transformer} -- selects $k$ tokens given a query and a table and a task-specific transformer solves the task based on these tokens. Decomposing the problem into two simpler tasks imposes additional structure that makes training more efficient: The first model is shallow, allowing the use of long input sequences at moderate cost, and the second model is deeper and uses the shortened input that solves the task. The combined model achieves a better efficiency-accuracy trade-off.

The pruning transformer is based on the \tapas QA model \cite{herzig-etal-2020-tapas}. \tapas answers questions by selecting tokens from a given table. This problem is quite similar to the pruning task.
The second transformer is a task-specific model adapted for each task to solve: We use another \tapas QA model for QA and a classification model ~\cite{eisenschlos2020understanding} for entailment.
In Section~\ref{sec:pruning}, we explain how we jointly learn both models by incorporating the pruning scores into the attention mechanism.

$DoT$ achieves a better trade-off between efficiency and accuracy on three datasets. We show that the pruning transformer selects relevant tokens, resulting in higher accuracy for longer input sequences.
We study the meaning of relevant tokens and show that the selection is deeply linked to solving the main task by studying the answer token scores. 
We open source the code in \url{http://github.com/google-research/tapas}.

\section{The $DoT$ Model}
\label{sec:pruning}

As show in Figure~\ref{fig:pruning_model}, the double transformer $DoT$ is composed of two transformers:
the \emph{pruning transformer} selects the most relevant $k$ tokens followed by a task-specific model that operates on the selected tokens to solve the task. The two transformers are learned jointly. $DoT$ loss is detailed in Appendix~\ref{app:loss}.
We explore learning the pruning model using an additional loss in Appendix~\ref{app:joint_learning}.

\begin{figure}
    \centering
    \includegraphics[width=0.65\linewidth]{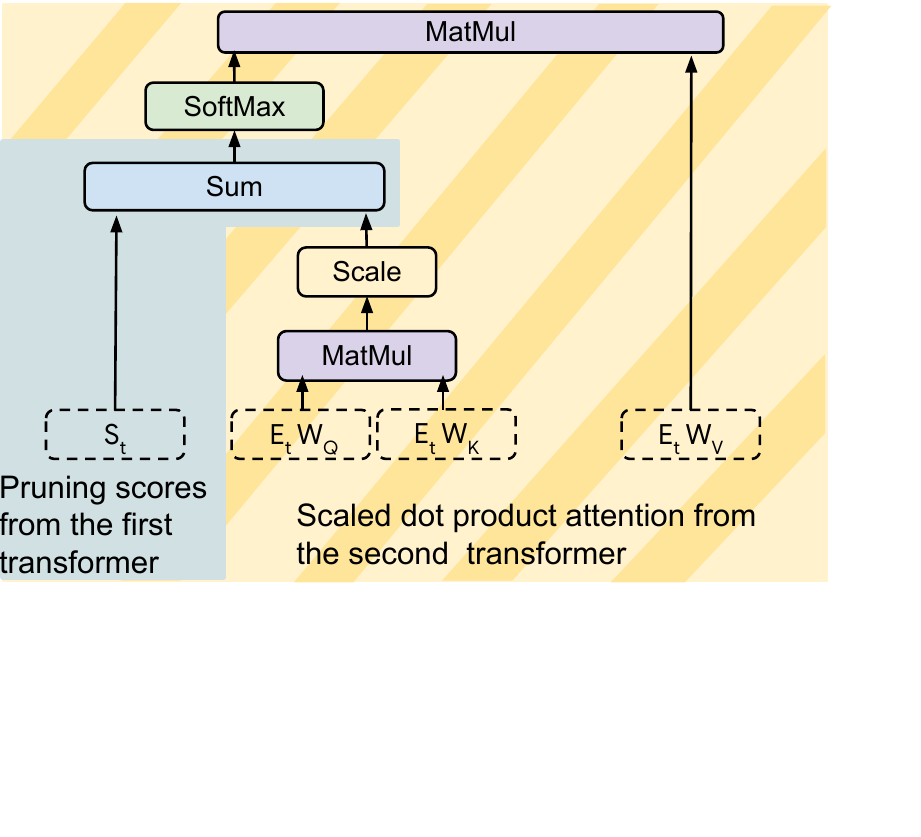}
    \vspace{-6.5ex} 
    \caption{Scaled dot product attention of the task model. We change the attention architecture~\cite{vaswani2017attention} -- the dashed bloc -- by adding the pruning scores -- the solid bloc. The pruning scores affect the task model's attention in all layers. This  enables  back  propagation  for  both  models based on a single loss.}
    \label{fig:attention}
\end{figure}

Let $q$ be the query (or statement) and $T$ the table. The transformer takes as input the embedding $E = [E_{[CLS]}; E_q; E_{[SEP]}; E_T ]$, composed of the query and table embeddings. The pruning transformer computes the probability $P(t |q,T)$ of the token $t$ being relevant to solve the example. We derive the pruning score $s_t = \log(P(t |q,T))$ and keep the top-$k$ tokens.
The pruning scores are then passed to the task transformer as shown in Figure~\ref{fig:attention}.

To enable the joint learning, we change the attention scores of the task model. For a normal transformer \cite{vaswani2017attention}, given the input embedding $E_t$ at position $t$, for each layer and attention head, the self-attention output is given by a linear combination of the value vector projections using the attention matrix.

Each row of the attention matrix is obtained by a softmax on the attention scores $z^{<t,t'>}$ given by

\vspace{-1.5ex}
\begin{equation}
\resizebox{0.3\vsize}{!}{
$z^{<t,t'>}=\frac{ E_t W_Q^{\top}  (E_{t'} W_K^{\top})^{\top}}{\sqrt{d_k}}
$
}
\end{equation}
where $W_Q$ and $W_K$
represent the query and key % and values
projections for that layer and head.
In our task model we add a negative bias term and replace this equation with
\vspace{-1.ex}
\begin{equation}
\resizebox{0.2\vsize}{!}{
  $  z^{<t, t'>|s_t} = z^{<t, t'>} + s_t $
}
\vspace{-1.0ex}
\end{equation}
Thus, the attention scores provide a notion of token \emph{relevance} -- detailed in Appendix~\ref{app:appendix-attention} -- and enable end-to-end learning of both models, letting $DoT$ define the top-$K$ tokens.

Unlike previous soft-masking methods~\cite{bastings-etal-2019-interpretable, de-cao-etal-2020-decisions}, ours coincides exactly with removing the input token $t$ when $P(t|q, T) \to 0$.
We prove this formally in Appendix~\ref{app:appendix-theorem}.

We explore two different pruning strategies: token selection defined as discussed above and column selection where we average all token scores in each column.
\section{Experimental Setup}
\label{sec:set-up}
We compare our approach against models using heuristic pruning.\\
\textbf{Cell concatenation ($CC$)} The \tapas model uses a default heuristic to limit the input tokens. The objective of the algorithm is to fit an equal number of tokens for each cell. This is done by first selecting the first token from each cell, then the second and so on until the desired limit is reached.\\
\textbf{Heuristic exact match ($HEM$)} \cite{eisenschlos2020understanding}. This method scores the columns based on their similarity to the question, where similarity is defined by token overlap.

We introduce a notation to clarify the setup:
$DoT(1^{st}_{type}\xrightarrow[]{top\mbox{-}k}2^{nd}_{type})$.
The $type$ correspond to the model size: small ($s$), medium ($m$) or large ($l$) as defined in \citet{2019arXiv190808962T}. For example, $CC \xrightarrow[]{1024} DoT(s \xrightarrow[]{256} l)$ denotes a $CC$ pre-processing to select $1024$ tokens passed to the $DoT$ model: one small pruning model that selects $256$ tokens and feeds them into a large task model.

Baselines and $DoT$ (hyper-parameters in Appendix~\ref{app:appendix-hyper-parameters}) are initialized from models pre-trained with a MASK-LM task, the intermediate pre-training data \cite{eisenschlos2020understanding} and following \citet{herzig-etal-2020-tapas} on SQA~\cite{iyyer-etal-2017-search}. The $DoT$ transformers' complexity -- detailed in Appendix~\ref{app:models-complexity} -- is similar to a normal transformer where only some constants are changed.

We evaluate $DoT$ on three datasets.\\
\textbf{\wikisql} ~\cite{DBLP:journals/corr/abs-1709-00103} is a corpus of $80,654$ questions with SQL queries, related to $24,241$ Wikipedia tables.
Here we train and test in the weakly-supervised setting where the answer to the question is the result of the SQL applied to the table.
The metric we use is denotation accuracy.\\
\textbf{\wtq} ~\cite{pasupat-liang-2015-compositional} consists of $22,033$ question-answer pairs on $2,108$ Wikipedia tables.
The questions are complex and often require comparisons, superlatives or aggregation.
The metric we use is the denotation accuracy as computed by the official evaluation script.\\
\textbf{\tabfact} ~\cite{2019TabFactA} contains $118$K statements about $16$K Wikipedia tables, labeled as either entailed or refuted.
The dataset requires both linguistic reasoning and symbolic reasoning with operations such as comparison, filtering or counting. We use the classification accuracy as metric.

In all our experiments we report results for $DoT$ using token selection for \wikisql and \tabfact and a column selection for \wtq.
\section{Results}
\label{sec:results}

\begin{table*}
\centering
\resizebox{0.75\textwidth}{!}{
\begin{tabular}{l*{3}{|lll}}
% \bottomrule
\textbf{Dataset} & \multicolumn{3}{|c}{\textbf{\wikisql}} & \multicolumn{3}{|c}{\textbf{\tabfact}} & \multicolumn{3}{|c}{\textbf{\wtq}} \\
\bottomrule
\textbf{Model}                                             & \textbf{test accuracy} & \textbf{Best}     & \textbf{NPE/s} &
                                                             \textbf{test accuracy} & \textbf{Best}     & \textbf{NPE/s} &
                                                             \textbf{test accuracy} & \textbf{Best}     & \textbf{NPE/s} \\
\bottomrule
\sota                                                      & \boldmath$83.9$        &                   &                &
                                                             \boldmath$81.0$        &                   &                &
                                                             \boldmath$51.8 \pm 0.6$& \boldmath$52.3$   &\\
\bottomrule
$\text{\ \ \ \ } CC \text{ }\xrightarrow[]{256}\tapas(l)$&  $76.4 \pm 0.3$          &  $77.15$          & $1870$         &
                                                            $75.1 \pm 0.3$          &  $76.13$          & $1900$       &
                                                            $44.8 \pm 0.5$          &  $45.47$          & $1900$ \\
$\text{\ \ \ \ } CC \text{ }\xrightarrow[]{512} \tapas(l)$ & $83.6 \pm 0.1$          & $83.65$          & $800$          &
                                                             $81.3 \pm 0.2$          & $81.60$          & $870$          &
                                                             $52.2 \pm 0.5$          & $52.74$          & $810$ \\
$\text{\ \ \ \ } CC \xrightarrow[]{1024}\tapas(l)$         & \boldmath$86.0 \pm 0.3$ & \boldmath$86.6$  & \boldmath$270$ &
                                                                      $81.6 \pm 0.1$ &          $81.64$ &          $300$ &
                                                             $53.9 \pm 0.2$          & $54.30$          & $270$ \\
\bottomrule
$\text{\ \ \ \ } CC \xrightarrow[]{1024}DoT(s\text{ } \xrightarrow[]{256} l)$ &  $74.2 \pm 3.6$           & $84.27$           & $1250$ &
                                                                                   $81.0 \pm 0.1$           & $81.17$           & $1300$ &
                                                                                   $48.1 \pm 2.4$           & $49.47$          & $1250$ \\
$\text{\ \ \ \ } CC \xrightarrow[]{1024}DoT(m \xrightarrow[]{256} l)$         &  $83.6 \pm 0.5$           & $84.67$           & $950$    &
                                                                                   $79.0 \pm 0.5$           & $81.28$           & $930$ 
                                                                              & \boldmath$50.1 \pm 0.5$   & \boldmath$50.14$  & \boldmath$950$\\ 
\bottomrule
\bottomrule
$HEM \xrightarrow[]{256} \tapas(l)$        &               $ 77.4 \pm 0.3 $ &  $77.97$ & $1870$  &
                                                           $ 75.5 \pm 0.2 $ &  $75.80$ & $1900$  &
                                                           $ 47.3 \pm 0.1 $ &  $47.70$ & $1900$  \\
$HEM \text{ }\xrightarrow[]{512} \tapas(l)$ &                $83.8 \pm 0.4$          & $84.75$          & $800$          &
                                                            \boldmath$82.0 \pm 0.3$  & \boldmath$82.07$ & \boldmath$870$ &
                                                            $52.7 \pm 0.4$           & $53.61$          & $810$ \\
$HEM \xrightarrow[]{1024} \tapas(l)$        &               $85.9 \pm 0.0$           & $85.94$          & $270$ &
                                                            $80.6 \pm 0.0$           & $80.6$           & $300$          &
                                                            \boldmath$54.0 \pm 0.9$  & \boldmath$54.93$ & \boldmath$270$ \\
\bottomrule
$HEM \xrightarrow[]{1024}DoT(s\text{ } \xrightarrow[]{256} l)$                &  \boldmath$85.3 \pm 0.4$  & \boldmath$85.76$  & \boldmath$1250$   &
                                                                                 \boldmath$81.6 \pm 0.3$  & \boldmath$81.74$  & \boldmath$1300$   &
                                                                                    $40.9 \pm 0.2$        &      $41.23$          & $1250$\\
$HEM \xrightarrow[]{1024}DoT(m \xrightarrow[]{256} l)$                        &  \boldmath$85.5 \pm 0.2$  & \boldmath$85.82$  & \boldmath$950$   &
                                                                                            $81.8 \pm 0.0$  & $81.94$           &  $930$   &
                                                                                            $40.1 \pm 2.4$  &   $49.13$         & $950$
% \bottomrule
\end{tabular}
}
% \vspace{-1ex}
\caption{Efficiency accuracy trade-off. We run $DoT$ with token pruning for \wikisql and \tabfact and column pruning for \wtq. The \sota (detailed in Appendix~\ref{app:sota}) corresponds to the models of \citet{min-etal-2019-discrete} for \wikisql,  \citet{eisenschlos2020understanding} for \tabfact and \citet{yin2020tabert} for \wtq. For each dataset, the \sota, the best baseline model on accuracy, and the $DoT$ models that reach the best accuracy efficiency trade-off are highlighted.}
\label{tab:results_efficency_accuracy}
\end{table*}

The baseline \tapas model outperforms the previous \sota on all datasets (Table~\ref{tab:results_efficency_accuracy}):
$+2.1$ for \wikisql  ($CC \xrightarrow[]{1024} \tapas(l)$),
$+1.07$ for \tabfact ($HEM \xrightarrow[]{512} \tapas(l)$),
and  $+1.3$ for \wtq ($HEM \xrightarrow[]{1024} \tapas(l)$).
\paragraph{Efficiency accuracy trade-off}
Table~\ref{tab:results_efficency_accuracy} reports the accuracy test results along with the average number of processed examples per second $NPE/s$ computed at training time. Using $HEM$ as pre-processing step improves $DoT$ models compared to $CC$ for both \wikisql and \tabfact.
$DoT(m)$ and $DoT(s)$ reach better efficiency accuracy trade-off for \wikisql: with a small drop of accuracy by $0.4\%$ (respectively $0.7\%$), they are $3.5$ (respectively $4.6$) times faster than the best baseline. For \tabfact dataset, $DoT$ is compared to a faster baseline than the one used for \wikisql as it takes only $512$ input tokens instead of $1024$. $DoT(s)$ still achieves a good trade-off: with a decrease of $0.4\%$ of accuracy it is $1.5$ times faster. Unlike the previous datasets, \wtq is a harder task to solve and requires passing more data. By restricting $DoT(m)$ to select only $256$ tokens we decrease the accuracy by a bigger drop $3.9\%$ to be $3.5$ times faster compared to $HEM\xrightarrow[]{1024} \tapas(l)$.

\paragraph{Small task models}
The previous results, raise the question of whether a smaller task model can reach a similar accuracy. To answer this question, we compare $\xrightarrow[]{1024}DoT(s\text{ } \xrightarrow[]{256} l)$ to $\xrightarrow[]{1024}\mbox{TAPAS}(s)$ and $\xrightarrow[]{256} \mbox{TAPAS}(l)$ in Table~\ref{tab:compare-to-smaller-transformers}. $DoT$ outperforms the smaller models showing the importance of using both transformers.
\begin{table}
\centering
\resizebox{\columnwidth}{!}{
\begin{tabular}{l*{3}{|ll}}
\textbf{Dataset} & \multicolumn{2}{|c}{\textbf{\wikisql}} & \multicolumn{2}{|c}{\textbf{\tabfact}} & \multicolumn{2}{|c}{\textbf{\wtq}} \\
\bottomrule
\textbf{Model}      & \textbf{test accuracy}    & \textbf{NPE/s} &
                     \textbf{test accuracy}     & \textbf{NPE/s} &
                     \textbf{test accuracy}     & \textbf{NPE/s} \\
\bottomrule
$CC\xrightarrow[]{1024} DoT(m \xrightarrow[]{256} l)$ & \boldmath$83.6 \pm 0.5$ & $950 $ & \boldmath$79.0 \pm 0.9$   & $930 $ & \boldmath$50.1 \pm 0.5$ & $950 $\\
$CC\> \> \xrightarrow[]{256} \mbox{TAPAS}(l)$         & $76.4 \pm 0.3$          & $1870$ & $75.1 \pm 0.3$            & $1900$ & $44.8 \pm 0.5$          & $1900$\\
$CC\xrightarrow[]{1024} \mbox{TAPAS}(m)$              & $81.6 \pm 0.2$          & $2050$ & $75.1 \pm 0.2$            & $2300$ & $42.9 \pm 0.3$          & $2020$\\
\end{tabular}
}
\caption{Comparing $DoT$ to smaller models similar to each of its two transformers.}
\label{tab:compare-to-smaller-transformers}
\end{table}
\section{Analysis}
\label{sec:analysis}

\begin{table}
\resizebox{\columnwidth}{!}{
\begin{tabular}{ll*{3}{l}}
\textbf{Bucket} & \textbf{Model}  & \textbf{\wikisql} & \textbf{\tabfact} & \textbf{\wtq} \\
\bottomrule
$ > 1024$ &$ \text{ } CC \> \>  \xrightarrow[]{512} \tapas(l)$      & \boldmath$24.3 \pm 0.1 $ &
                                                            \boldmath$56.8 \pm 2.2$  &
                                                            \boldmath$18.8 \pm 0.9$  \\ 
&$ \text{ } CC \> \> \xrightarrow[]{256} \tapas(l)$       & $ 5.8 \pm 0.2$ &
                                                            $ 9.9 \pm 1.5$   &
                                                            $6.9 \pm 0.0$   \\
&$\text{ } CC \xrightarrow[]{1024} DoT(m\xrightarrow[]{256}l)$& \boldmath$ 40.1 \pm 4.9 $&
                                                            \boldmath$69.1 \pm 2.5$ &
                                                            \boldmath$23.8 \pm 0.5$  \\
\bottomrule
$[512, 1024]$ &$  \text{ } CC  \> \> \xrightarrow[]{512}\tapas(l)$ &         \boldmath$73.6 \pm 0.1 $&
                                                                \boldmath$73.0 \pm 0.3$ &
                                                                \boldmath$42.7 \pm 0.6$ \\
& $  \text{ } CC\> \> \xrightarrow[]{256}\tapas(l)$&            $40.9 \pm 0.3$ &
                                                                $43.9 \pm 0.4$&
                                                                $18.6 \pm 0.1$ \\
& $ \text{ } CC \xrightarrow[]{1024} DoT(m\xrightarrow[]{256}l)$&   \boldmath$72.9 \pm 1.8$ &
                                                                \boldmath$74.7 \pm 0.7$ &
                                                                \boldmath$39.1 \pm 0.6$
\end{tabular}
}
\caption{The denotation accuracy for test, computed over bucketized datasets per sequence length. The pruning transformer prunes efficiently --two times better: $DoT(\xrightarrow[]{256}l)$ reaches accuracy close and higher than using $CC\xrightarrow[]{512}$ heuristic with $\tapas(l)$.
}
\label{tab:results-buckets}
\end{table}

\paragraph{Accuracy for long input sequences}
To study the long inputs, we bucketize the datasets per example input length. We compare $DoT(m\xrightarrow[]{256}l)$ to different $CC\xrightarrow[]{.}\tapas(l)$ models in Table~\ref{tab:results-buckets}. For the bucket $> 1024$ the $DoT$ model outperforms the $256$ and $512$ length baselines for all tasks. This indicates that the pruning model extracts two times more relevant tokens than the heuristic $CC$.

For the bucket $[512, 1024]$, we expect all models to reach a higher accuracy, as we expect lower loss of context than for the bucket $> 1024$ when applying $CC$. The results shows that $DoT$ gives a similar accuracy to $\xrightarrow[]{512} \tapas$ for \wikisql and \tabfact -- in the margin error -- and a slightly lower accuracy for \wtq: The pruning transformer selects only $256$ top-$K$ tokens compared to  $\xrightarrow[]{512} \tapas$ that selects twice more. Thus, the task-specific transformer has access to less tokens, therefore to possibly less context that can lead to an accuracy drop. This drop is small compared to $\xrightarrow[]{256} \tapas$ baseline drop. $DoT$ still outperforms $\xrightarrow[]{256} \tapas$ for all datasets.

\begin{figure}
    \centering
    \scalebox{1.}{
    \includegraphics[width=1.\linewidth]{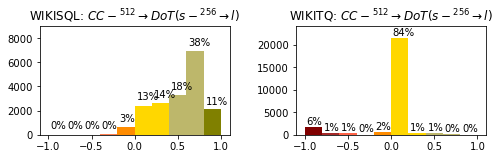}}
    \vspace{-5ex}
    \caption{Distribution of the answer token scores minus the average scores of the top-$K$ tokens. The difference is larger when the pruning transformer attributes a higher score to the answer tokens.}
    \label{fig:analysis_scores_dist}
\end{figure}

\paragraph{Pruning relevant tokens}
We inspect the pruning transformer on the \wikisql and \wtq datasets, where the set of answer tokens is given. We compute the difference between the answer token scores and the average scores of the top-$K$ tokens, and report the distribution in Figure~\ref{fig:analysis_scores_dist}. The pruning transformer tends to attribute high scores to the answer tokens, suggesting that it learns to answer the downstream question -- a positive difference -- especially for \wikisql. The difference is lower for \wtq as it is a harder task: The set of answer tokens is larger, especially for aggregation, making their scores closer to the average.

\paragraph{Pruning transformer depth} We study the pruning transformer complexity impact on the efficiency accuracy trade-off. Figure~\ref{fig:efficiency-accuracy-trade-off} compares the results of medium, small and mini models -- complexity in Appendix~\ref{app:models-complexity}. For all datasets the $mini$ model drops drastically the accuracy. The pruning transformer must be deep enough to learn the top-$K$ tokens and attribute token scores that can be used by the task-specific transformer.
For both \wikisql and \tabfact the small model reaches a better accuracy efficiency trade-off: Using a small instead of medium -- $4$ hidden layers instead of $8$ -- drops the accuracy by less than $0.4\%$ -- in the margin error -- while accelerating the model times $1.3$. In other words there is no gain of using a more complex model to select the top-$K$ tokens especially when we restrict $K$ to $256$.

Restricting $K$ can lead to a drop in the accuracy. Even by increasing the pruning complexity, $DoT$ cannot recover the full drop. This is the case of \wtq. This dataset is more complex, it requires more reasoning including operation to run over multiple cells in one column. Thus selecting the top $256$ tokens is a harder task compared to previous detests. We reduce the task complexity by using column selection instead of token selection. For this dataset using medium pruning transformer, $DoT(m)$ reaches a better accuracy efficiency trade-off: $2$ points higher in accuracy compared to using a small transformer.  

\begin{figure*}[ht]
    \centering
    \scalebox{1.0}{
    \includegraphics[width=1.0\textwidth]{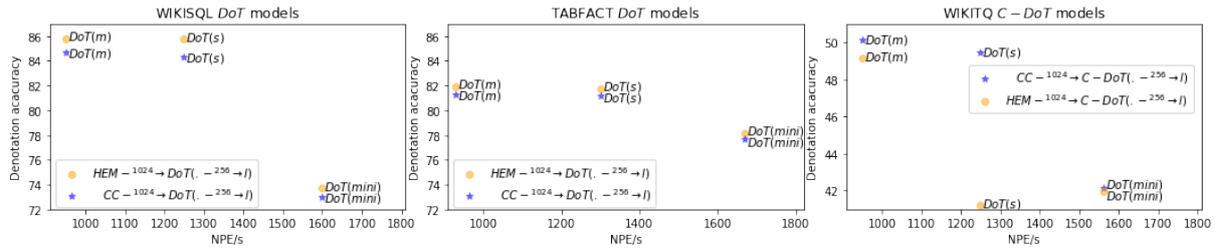}}
    \vspace{-4ex}
    \caption{$DoT$ models efficiency accuracy trade-off. The $DoT$ models are displayed according to their accuracy in function of the average number of processed examples par second. The models are faster being closer to the right side of the figures and have higher accuracy being closer to the top. This figure compares the efficiency accuracy trade-off of using different pruning transformers -- medium, small and mini -- and study the impact of $HEM$ and $CC$ on $DoT$. We use token selection for both \wikisql and \tabfact and column selection for \wtq.}
    \label{fig:efficiency-accuracy-trade-off}
\end{figure*}

\paragraph{Effects of $HEM$ and $CC$ on $DoT$}
Table~\ref{tab:results_efficency_accuracy} and Figure~\ref{fig:efficiency-accuracy-trade-off} compare the effect of using $HEM$ and $CC$ on $DoT$ models. As both heuristics are applied in the pre-processing step, using $HEM$ or $CC$ along with a similar $DoT$ model, doesn't change the average number of processed examples per second $NPE/s$ computed over the training step.
For both \wikisql and \tabfact we use a token based selection to select the top-$K$ tokens. Combining the token based strategy with $HEM$, outperforms on accuracy the token pruning $DoT$ combined with $CC$. For \wtq, the top-$K$ pruning is a column based selection. Unlike the token selection the column pruning combined with $HEM$ gives a lower accuracy.
\section{Related work}

\paragraph{Efficient Transformers} Improving the computational efficiency of transformer models, especially for serving, is an active research topic. Proposed approaches fall into four categories. The first is to use knowledge distillation, either during the pre-training phase~\cite{sanh2020distilbert}, or for building task-specific models~\cite{sun2019patient}, or for both~\cite{jiao2019tinybert}. The second category is to use quantization-aware training during the fine-tuning phase of \bert models, such as~\cite{zafrir2019q8bert}. The third category is to modify the transformer architecture to improve the dependence on the sequence length~\cite{choromanski2020rethinking,wang2020linformer}. The fourth category is to use pruning strategies such as \citet{mccarley2020structured}, who studied structured pruning to reduce the number of parameters in each transformer layer, and \citet{fan2020reducing} who used structured dropout to reduce transformer depth at inference time. Our method most closely resembles the last category, but we focus our efforts on shrinking the sequence length of the input instead of model weights. \citet{eisenschlos2020understanding} explore heuristic methods  based on lexical overlap and apply it to tasks involving tabular data, as we do, but our algorithm is learned end-to-end and more general in nature.
\paragraph{Interpretable NLP}
Another related line of work attempts to 
interpret neural networks by searching for \emph{rationales} \cite{lei-etal-2016-rationalizing}, which are a subset of words in the input text that serve as a justification for the prediction. \citet{lei-etal-2016-rationalizing} learn the rationale as a latent discrete variable inside a computation graph with the \abr{reinforce} method~\cite{reinforce}. 
\citet{bastings-etal-2019-interpretable} propose instead using stochastic computation nodes and continuous relaxations
\cite{maddison2017concrete}, % gumbeltrick
based on re-parametrization \cite{DBLP:journals/corr/KingmaW13} to approximate the discrete choice of a rationale from an input text, before using it as input for a classifier. 
Partially masked tokens are then replaced at the input embedding layer by some linear interpolation.
We rely on a soft attention mask instead as a way to partially reduce the information coming from some tokens during training.
To the best of our knowledge these methods have not been investigated in the context of semi-structured data such as tables or evaluated with a focus on efficiency.
\section{Conclusion}
We introduced double transformer ($DoT$) where an additional small model prunes the input of a larger second model.
This accelerates the training and inference time at a low drop in accuracy. 
As future work we will explore hierarchical pruning and adapt $DoT$ to other semi-structured NLP tasks.

\section*{Acknowledgments}
We thank Yasemin Altun, William Cohen, Slav Petrov and the anonymous reviewers for their constructive feedback, comments and suggestions.
\clearpage
\bibliographystyle{style/acl_natbib}
\bibliography{acl2020.bib}

\clearpage
\appendix
\appendix\section*{Appendix}
\label{app:appendix}
\section{$DoT$ model}
\subsection{Attention scores defines the meaning of relevant tokens.}
\label{app:appendix-attention}
We study, the updates of the pruning scores according to the attention scores needs. We note the set of relevant tokens $R$. The output probability given by the pruning transformer is in $(0,1)$ making $s_t$ in $(-\infty, 0)$. Lets suppose that the token $t$ is not needed to answer the question, then the attention scores are decreased $z^{<i,t,t'>|s_t} \to \Vec{-\infty}$ for all the tokens $t' \in R$ for all the layers $i$.
The model updates both parts of $z^{<i,t,t'>}$ making $s_t$ converging to $-\infty$, then $\lim_{s_t\to-\infty} z^{<i,t,t'>|s_t} = \Vec{-\infty}$. Thus, the meaning of relevant token is defined by the attention scores updates: The pruning scores decreases for non relevant tokens and increase for relevant ones.

\subsection{$DoT$ loss}
\label{app:loss}
The $DoT$ loss is similar to \tapas model loss -- noted as $J_{SA} = J_{aggr} + \beta J_{scalar}$ in \cite{herzig-etal-2020-tapas} -- computed over the task-specific transformer where the attention scores are modified. More precisely, we modify only the scalar loss of the task specific model $J_{scalar}$. We incorporate the pruning scores $S = \{s_t \forall t \in T_{top_k=256}\}$, and we note $J_{scalar|S}$. The $DoT$ loss is then compute only over the top-$K$ tokens: $J_{DoT} = J_{aggr} + \beta J_{scalar|S}$.

For \tabfact dataset, \citet{eisenschlos2020understanding} modified the \tapas loss -- used for QA tasks -- to adapt it to the entailment task: Aggregation is not used, instead, one hidden layer is added as output of the [CLS] token to compute the probability of Entailment. We use a similar loss for \tabfact where the attention scores are modified.

\subsection{Feed-forward pass: Safe use of shorter inputs for the task-specific transformer}
\label{app:appendix-theorem}
The top-$K$ selection enables the use of shorter inputs for the task-specific.
We prove that using input length equal to K is equivalent to using input length higher than K, without any loss of context.
Note that the pruning scores are the same for both inputs, where the top-$K$ are scored non-zero and we impose the other tokens to be scored zero. 
\begin{theorem}
\label{th:attention}
Given a transformer and a set of tokens as input $I$. Let $t$ be one of the input tokens $t \in I$. If the transformer verifies the following conditions, that holds for all layers $i$.
\begin{enumerate}
    \item \label{th:condition-1}  $\forall t' \in I$ that attends to $t$, , $z^{<i,t,t'>} = \Vec{-\infty}$.
    \item \label{th:condition-2} For $t$ attends to any $ t' \in I$, $z^{<i,t',t>} = \Vec{-\infty}$.
\end{enumerate}
Then applying this transformer on $I$ is equivalent to applying it on $I-\{t\}$
\end{theorem}

\begin{proof}
We look at the different use cases.

$\forall i$ layers, any token $ t' \in I-\{t\}$ attending to any token $t'' \in I-\{t\}$: the soft-max scores $a^{<i,t',t''>} $ have the same formula using $I$ or $I-\{t\}$ as input.

Lets fix $t'=t$. The token $t$ attending to any token $t'' \in I$: The first condition~\ref{th:condition-1} gives $\forall t''$ that attends to $t$, $z^{<i,t,t''>} = \Vec{-\infty}$. That follows $exp(z^{<i,t,t''>}) = \Vec{0}$ then $a^{<i,t,t''>}=\Vec{0}$.

Similarly, if $t'=t$. Any token $t'' \in I$ attending to $t$: The second condition~\ref{th:condition-2} gives $\forall t'$ that attends to $t$,  $z^{<i,t,t''>} = \Vec{-\infty}$. That follows $exp(z^{<i,t',t>}) = \Vec{0}$ then $a^{<i,t',t>}=\Vec{0}$.
\end{proof}

\begin{remark}
Given a transformer and a set of tokens as input $I$. Let $t$ be one of the input tokens $t \in I$ with $t$ is not selected by the pruning transformer scored zero -- not the first-k tokens. Using $DoT$, $s_t = -\infty$. That follows $z^{<i,t,t'>} = \Vec{-\infty}$.

The case $t'=t$, for any token $t'' \in I$ attending to $t$ we have: $\forall i \neq 0$, the input $E_t = \sum_{t'' \in I} a^{<i,t,t''>} $. As $z^{<i,t,t'>} = \Vec{-\infty}$, $E_t = \Vec{0}$, $E_t $ zero out all the variables making $exp(z^{<i,t',t>})$ a constant and $a^{<i,t',t>}$ independent of $t'$. This is equivalent to $\forall i \neq 0$, $t$ doesn't attend to any $t' \in I$.

Only for the first layer $i = 0$, we add an approximation to drop the attention ($t$ attending to $t' \in I$). We consider the impact of $t$ on the full attention is small as we stuck multiple layers. We experimented with a task-specific model with a big input length $>k $ and compare it to a task-specific model with input length $=k$. The two models gives similar accuracy. In our experiment we report only the results for the model with input length $=k$.

This makes the attention scores similar to the ones computed over $t \notin I$.
\end{remark}

\section{Experiments}
In all the experiment we report the median accuracy and the error margin computed over 3 runs. We estimate the error margin as half the inter quartile range, that is half the difference between the $25^{th}$ and $75^{th}$ percentiles.
\subsection{Models hyper-parameters}
\label{app:appendix-hyper-parameters}
We do not perform hyper-parameters search for $DoT$ models we use the same as \tapas baselines. For \wikisql and \wtq we use the same hyper-parameters as the one used by \cite{herzig-etal-2020-tapas} and for \tabfact the one used by \cite{eisenschlos2020understanding}. Baselines and $DoT$ are initialized from models pre-trained with a MASK-LM task and on SQA\cite{iyyer-etal-2017-search} following \citet{herzig-etal-2020-tapas}.

We report the models hyper parameters used for \tapas baselines and $DoT$ in Table~\ref{tab:hyper-parameters-datasets}. The hyper-parameters are fixed independently of the pre-processing step or the input size: For all the pre-processing input lengths -- $\{256,512,1024\}$--, for both $CC$ and $HEM$ we use the same hyper-parameters.
Additionally, we use an Adam optimizer with weight decay for all the baselines and $DoT$ models --the same configuration as \bert.

\begin{table}
\centering
\resizebox{\columnwidth}{!}{
\begin{tabular}{llllll}
\textbf{Dataset}  & \textbf{lr} & \textbf{$\rho$} & \textbf{hidden dropout} & \textbf{attention dropout} &  \textbf{num steps}\\
\bottomrule
$\wikisql$ & $6e^{-5}$   & $0.14$ & $0.1$  & $0.1$ & $50, 000$ \\
$\tabfact$ & $2e^{-5}$   & $0.05$ & $0.07$ & $0.0$ & $80, 000$\\
$\wtq$     & $1.9e^{-5}$ & $0.19$ & $0.1$  & $0.1$ & $50, 000$
\end{tabular}
}
\vspace{-1ex}
\caption{Hyper-parameters used per dataset. Reports the learning rate (lr), the warmup ratio ($\rho$), the hidden dropout, the attention dropout and the number of training steps (num steps) used for each dataset. These hyper-parameters are the same for all the baselines and $DoT$ models.}
\label{tab:hyper-parameters-datasets}
\end{table}

\subsection{\sota}
\label{app:sota}
We report \sota for the three datasets in Table~\ref{tab:sota}.
\begin{table}
\centering
\subcaptionbox{\sota \wikisql\label{tab:sota_wikisql}}
{\resizebox{\columnwidth}{!}
{\begin{tabular}{ll}
    \textbf{Model} & \textbf{test accuracy} \\ 
    \midrule
    \cite{agarwal2019learning} MeRL & $74.8 \pm 0.2$ \\
    \cite{liang2019memory} MAPO (ensemble of 10) & $ 74.9$\\
    \cite{wang-etal-2019-learning-semantic} & $79.3$ \\
    $CC \xrightarrow[]{512}\tapas(l)$\cite{herzig-etal-2020-tapas} &$83.6$ \\
    \textbf{\cite{min-etal-2019-discrete}} & \boldmath$83.9$ \\
    \bottomrule
\end{tabular}}}
\subcaptionbox{\sota \tabfact \label{tab:sota_tabfact}} 
{\resizebox{\columnwidth}{!}{
    \begin{tabular}{ll}
    \textbf{Model} & \textbf{test accuracy} \\ 
    \midrule
    \cite{zhong-etal-2020-logicalfactchecker} LFC (LPA) & $71.6$ \\
    \cite{zhong-etal-2020-logicalfactchecker} LFC (Seq2Action) & $71.7$ \\
    \cite{shi-etal-2020-learn} HeterTFV & $72.3$ \\
    \cite{zhang2020table} SAT & $73.2$ \\
    \cite{yang2020program} ProgVGAT & $74.4$\\
    \boldmath$CC \xrightarrow[]{512}\tapas(l)$ \cite{eisenschlos2020understanding} & \boldmath$81.0$ \\
    \bottomrule
    \end{tabular}
    }
}
\subcaptionbox{\sota \wtq \label{tab:sota_wtq}} 
{\resizebox{\columnwidth}{!}{
    \begin{tabular}{ll}
    \textbf{Model} & \textbf{test accuracy} \\ 
    \midrule
    \cite{agarwal2019learning} MeRL & $44.1 \pm 0.2$ \\
    \cite{dasigi-etal-2019-iterative} Iterative Search (best) & $44.3$  \\ 
    \cite{wang-etal-2019-learning-semantic} &$44.5$\\
    \cite{liang2019memory} MAPO (ensembled-$10$) &$ 46.3$\\ 
    \cite{agarwal2019learning} MeRL ensemble of $10$ models & $46.9$ \\
    $CC \xrightarrow[]{512}\tapas(l)$\cite{herzig-etal-2020-tapas}& $48.8$ \\
    \textbf{\cite{yin2020tabert}} \boldmath$MAPO + TABERT(l)(K = 3)$ & \boldmath$51.8 \pm 0.6$ \\
    \bottomrule
    \end{tabular}}
}
\caption{\sota accuracy on test set.}
\label{tab:sota}
\end{table}

\subsection{Models complexity}
\label{app:models-complexity}
In all our experiments we use different transformer sizes called large, medium, small and mini. These models correspond to the \bert open sourced model sizes described in \citet{2019arXiv190808962T}. We report all models complexity in Table~\ref{tab:models_complexity}.
\begin{table}
\centering
\resizebox{0.8\columnwidth}{!}{
\begin{tabular}{lllll}
\textbf{Model} & \textbf{$\#L$}  & \textbf{$H$} & \textbf{$\#Hs$} &  \textbf{$Hi$} \\
\bottomrule
large  & $24$ & $1024$ & $16$ & $4096$  \\
medium & $8$  & $512$  & $8$  & $2048$  \\
small  & $4$  & $512$  & $8$  & $2048$  \\
mini   & $4$  & $256$  & $4$  & $1024$ 
\end{tabular}
}
\vspace{-1ex}
\caption{Models complexity with $\#L$ is the number of layers, $\#Hs$ the number of heads, $H$ the embedding size and $Hi$ the intermediate size.}
\label{tab:models_complexity}
% \vspace{-3ex}
\end{table}
The sequence length changes the total number of used parameters. The formula to count the number of parameters is given by Table~\ref{tab:params_count}.
\begin{table}
\resizebox{\columnwidth}{!}{
\begin{tabular}{lll}
\textbf{Num layers $\times$ Module} & \textbf{Tensor}  & \textbf{Shape} \\
\bottomrule
$1\times $Embedding    & embeddings.word\_embeddings                       & $[V, H]$\\
                        &embeddings.position\_embeddings                   & $[I, H]$\\
                        &embeddings.token\_type\_embeddings                & $[3, H]$\\
                        &                                                  & $ +[2, H]$\\
                        &                                                  & $+[10, H]$\\
                        &                                                  & $+4[256,H]$\\
                        &embeddings.LayerNorm                              & $[H] $\\
                        &                                                  & $+ [I]$\\
\bottomrule
$L\times $Transformer  & encoder.layer.0.attention.self.query.kernel       & $[I, H]$\\
                        &encoder.layer.0.attention.self.query.bias         & $[H]   $\\
                        &encoder.layer.0.attention.self.key.kernel         & $[I, H]$\\
                        &encoder.layer.0.attention.self.key.bias           & $[H]   $\\
                        &encoder.layer.0.attention.self.value.kernel       & $[I, H]$\\
                        &encoder.layer.0.attention.self.value.bias         & $[H]   $\\
                        &encoder.layer.0.attention.output.dense.kernel     & $[H, H]$\\
                        &encoder.layer.0.attention.output.dense.bias       & $[H]   $\\
                        &encoder.layer.0.attention.output.LayerNorm        & $[H] $\\
                        &                                                  & $+ [H]$\\
                        &encoder.layer.0.intermediate.dense.kernel         & $[H, Hi]$\\
                        &encoder.layer.0.intermediate.dense.bias           & $[Hi]  $\\
                        &encoder.layer.0.output.dense.kernel               & $[Hi, H]$\\
                        &encoder.layer.0.output.dense.bias                 & $[H]   $\\
                        &encoder.layer.0.output.LayerNorm                  & $[H] $\\
                        &                                                  & $+ [H]$\\
\bottomrule
$1\times $ Pooler      & pooler.dense.kernel                               & $[I, H]$\\
                        &pooler.dense.bias                                 & $[H]   $\\
\bottomrule
\end{tabular}
}
\vspace{-1ex}
\caption{Parameters counts. Let $H$ be the hidden embedding size, $L$ the number of layers, $Hi$ the intermediate size, $V=30522$ the vocabulary size and $I$ the input size. We report the used parameters based on tensors shape for \tapas models.}
\label{tab:params_count}
% \vspace{-3ex}
\end{table} 
The number of used parameters equals to $V \times H + (2+3L) I \times H + I + (256*4 + 17 + 9L)H + (1 + 2L \times H) \times Hi$. The number of parameters of each model is reported in Table~\ref{tab:all_models_parameters}

\begin{table}
\resizebox{\columnwidth}{!}{
\begin{tabular}{l*{1}{|lll}}
\textbf{Model} & \multicolumn{3}{c}{\textbf{Parameters count}}\\

\textbf{}  & \textbf{$CC\xrightarrow[]{256}$} & \textbf{$CC\xrightarrow[]{512}$} & \textbf{$CC\xrightarrow[]{1024}$}\\
\bottomrule
$ \mbox{TAPAS}(mini)$    & $ 11.1M$ & $ 12.0M$ & $ 13.8M$\\
$ \mbox{TAPAS}(s)$    & $ 26.4M$ & $ 28.2M$ & $ 31.9M$\\
$ \mbox{TAPAS}(m)$    & $ 36.3M$  & $ 39.7M$ & $ 46.6M$\\
$\tapas(l)$&  $253.2M$ & $ 272.6M$  &  $ 311.4 M$\\
$DoT(mini \xrightarrow[]{256} l)$ & $264.3 M$  & $265.3M$  & $267.1M$\\
$DoT(s \text{ } \xrightarrow[]{256} l)$ & $279.6 M$  & $ 281.5M$  & $285.1M$\\
$DoT(m \xrightarrow[]{256} l)$          & $ 289.6 M$  & $ 293M$ & $299.8M$ \\

\end{tabular}
}
\caption{The parameters count for the different models. $M$ refers to millions. The number of parameters is the same using $HEM$ or $CC$. The column based $DoT$ models have the same number of parameters than the token based $DoT$ models.}
\label{tab:all_models_parameters}
\end{table}
The number of parameters is not proportional to the computational time as multiple operations involves multiplying tensors of shapes $[I,H]\times [H,H]$.

\section{Analysis}
\label{app:analysis}
We report additional results for the analysis.
\subsection{Pruning transformer enables reaching high accuracy for long input sequences}
\label{app:buckets}
To study the model accuracy on different input sequence lengths, we bucketize the datasets. Table~\ref{tab:results-buckets-all} reports the accuracy results computed over the test set for all buckets. We use $DoT(m\xrightarrow[]{256}l)$ model for the three datasets, a token based pruning for both \wikisql and \tabfact and a column based pruning for \wikisql. For a length $> 1024$, the $DoT$ model outperforms the $256$ and $512$ length baselines for all tasks. For the bucket $[512, 1024]$, $DoT$ model gives close results to $512$ length baseline. This indicates that the pruning model extracts twice more relevant tokens than the heuristic $CC$. For smaller input lengths the baseline models outperform $DoT$. One cause could be the hyper-parameters tuning as we do not tune the hyper parameters for $DoT$.  
\begin{table}
\resizebox{\columnwidth}{!}{
\begin{tabular}{ll*{3}{l}}
\textbf{Bucket} & \textbf{Model}  & \textbf{\wikisql} & \textbf{\tabfact} & \textbf{\wtq} \\
\bottomrule
$ > 1024$ & $ \text{ } CC \xrightarrow[]{1024}\tapas(l)$ & $54.7 \pm 1.0 $ &
                                                            $74.1 \pm 1.5$ &
                                                            $30.7 \pm 0.9$  \\
&$ \text{ } CC \> \>  \xrightarrow[]{512} \tapas(l)$      & \boldmath$24.3 \pm 0.1 $ &
                                                            \boldmath$56.8 \pm 2.2$  &
                                                            \boldmath$18.8 \pm 0.9$  \\ 
&$ \text{ } CC \> \> \xrightarrow[]{256} \tapas(l)$       & $ 5.8 \pm 0.2$ &
                                                            $ 9.9 \pm 1.5$   &
                                                            $6.9 \pm 0.0$   \\
&$\text{ } CC \xrightarrow[]{1024} DoT(m\xrightarrow[]{256}l)$& \boldmath$ 40.1 \pm 4.9 $&
                                                            \boldmath$69.1 \pm 2.5$ &
                                                            \boldmath$23.8 \pm 0.5$  \\
\bottomrule
$[512, 1024]$ &$ \text{ } CC \xrightarrow[]{1024} \tapas(l)$ & $86.1 \pm 0.5$&
                                                                $78.0 \pm0.4$ &
                                                                $48.2 \pm 0.1$ \\
&$  \text{ } CC  \> \> \xrightarrow[]{512}\tapas(l)$ &         \boldmath$73.6 \pm 0.1 $&
                                                                \boldmath$73.0 \pm 0.3$ &
                                                                \boldmath$42.7 \pm 0.6$ \\
& $  \text{ } CC\> \> \xrightarrow[]{256}\tapas(l)$&            $40.9 \pm 0.3$ &
                                                                $43.9 \pm 0.4$&
                                                                $18.6 \pm 0.1$ \\
& $ \text{ } CC \xrightarrow[]{1024} DoT(m\xrightarrow[]{256}l)$&   \boldmath$72.9 \pm 1.8$ &
                                                                \boldmath$74.7 \pm 0.7$ &
                                                                \boldmath$39.1 \pm 0.6$\\
\bottomrule
$[256, 512]$ &$ \text{ } CC \xrightarrow[]{1024} \tapas(l)$ &   $87.0 \pm 0.0$&
                                                                $80.3 \pm 0.2$ &
                                                                $56.1 \pm 0.4$ \\
&$  \text{ } CC  \> \> \xrightarrow[]{512}\tapas(l)$ &          $88.2 \pm 0.1$&
                                                                $81.1 \pm 0.2$ &
                                                                $56.5 \pm 1.1$ \\
& $  \text{ } CC\> \> \xrightarrow[]{256}\tapas(l)$&            $78.5 \pm 0.7$ &
                                                                $71.4 \pm 0.3$&
                                                                $50.2 \pm 0.8$ \\
& $ \text{ } CC \xrightarrow[]{1024} DoT(m\xrightarrow[]{256}l)$&$86.4 \pm 0.7$ &
                                                                $78.4 \pm 0.5$ &
                                                                $53.1 \pm 0.5$\\
\bottomrule
$<256 $ &$ \text{ } CC \xrightarrow[]{1024} \tapas(l)$ &        $87.7 \pm 0.0$&
                                                                $82.2 \pm 0.1$ &
                                                                $58.7 \pm 0.2$ \\
&$  \text{ } CC  \> \> \xrightarrow[]{512}\tapas(l)$ &          $88.2 \pm 0.3$&
                                                                $83.2 \pm 0.0$ &
                                                                $60.2 \pm 1.0$ \\
& $  \text{ } CC\> \> \xrightarrow[]{256}\tapas(l)$&            $88.4 \pm 1.1$ &
                                                                $82.1 \pm 0.2$&
                                                                $59.1 \pm 0.7$ \\
& $ \text{ } CC \xrightarrow[]{1024} DoT(m\xrightarrow[]{256}l)$&$87.8 \pm 0.2$ &
                                                                $79.9 \pm 0.5$ &
                                                                $57.3 \pm 0.6$\\
\end{tabular}
}
\vspace{-1ex}
\caption{The denotation accuracy for test, computed over bucketized datasets per sequence length.}
\label{tab:results-buckets-all}
% \vspace{-3ex}
\end{table}

\subsection{Choice of joint learning: Is it better to impose the meaning of relevant tokens?}
\label{app:joint_learning}
According to the analysis done in Section~\ref{sec:analysis}, the pruning model --jointly learned-- is selecting the tokens to solve the main task. This raises a question of whether adding a pruning loss similar to the task-specific loss can improve the end-to-end accuracy. We not $J_{pruning-scalar}$ the pruning loss and $J_{task-specific-scalar}$ the task-specific loss. Both are similar to $J_{scalar}$ defined by \cite{herzig-etal-2020-tapas} where the attention scores are not affected by the pruning scores. We additionally not $J_{task-specific-scalar|S}$ the task specific loss affected by the set of pruning scores $S$.

We compare the joint learning model $J\mbox{-}DoT(.)$ -- defined in Appendix~\ref{app:loss} -- to a model learned using an additional pruning loss $P \mbox{-}DoT(.) = J_{aggr} + \beta (J_{task-specific-scalar} + J_{pruning-scalar})$, and another using both $PJ \mbox{-}DoT(.) = J_{aggr} + \beta (J_{task-specific-scalar|S} + J_{pruning-scalar})$. Table~\ref{tab:results-loss} shows that for both \wikisql and \wtq joint learning achieves higher accuracy for similar efficiency. For \tabfact the median is in the margin error but the best model using the joint learning outperforms the other learning strategies.  
\begin{table*}
\centering
\resizebox{\textwidth}{!}{
\begin{tabular}{l*{3}{lll}}
\textbf{Dataset} & \multicolumn{3}{c}{\textbf{\wikisql}} & \multicolumn{3}{c}{\textbf{\tabfact}} & \multicolumn{3}{c}{\textbf{\wtq}} \\
\bottomrule
\textbf{Model} & \textbf{test accuracy} & \textbf{Best} & \textbf{$NPE/s$}&  
                 \textbf{test accuracy} & \textbf{Best} & \textbf{$NPE/s$}&  
                 \textbf{test accuracy} & \textbf{Best} & \textbf{$NPE/s$} \\ 
\bottomrule
$CC \xrightarrow[]{1024}P\mbox{-}DoT(m \xrightarrow[]{256} l)$       &  $80.4 \pm 0.6$           & $82.11$          & $950$ &
                                                                        $79.7 \pm 0.5$           & $80.20$          & $930$ &
                                                                        $43.5 \pm 0.6$           & $44.54$          & $950$\\
$CC \xrightarrow[]{1024}PJ\mbox{-}DoT(m \xrightarrow[]{256} l)$      &  $82.9 \pm 0.6$           & $83.21$          & $950$ &
                                                                        $78.0 \pm 0.3$           & $78.41$          & $930$ &
                                                                        $46.4 \pm 0.8$           & $48.43$          & $950$\\
\boldmath$CC \xrightarrow[]{1024}J\mbox{-}DoT(m \xrightarrow[]{256} l)$& \boldmath$83.6 \pm 0.5$ & \boldmath$84.67$ & \boldmath$950$ &
                                                                         \boldmath$79.0 \pm 0.5$ & \boldmath$81.28$ & \boldmath$930$  &
                                                                         \boldmath$50.1 \pm 0.5$ & \boldmath$50.14$ & \boldmath$950$
\end{tabular}
}
\caption{Test denotation accuracy using different $DoT$ training losses. We compare a joint learning $J\mbox{-}DoT$ model -- that enables the back propagation of the pruning transformer by modifying the attention scores -- to two different $DoT$ models where we modify the learning strategy. $P\mbox{-}DoT$ disables the buck-propagation to the pruning transformer through the attention scores, instead it uses an additional pruning loss similar to the task-specific loss. The second strategy, $JP\mbox{-}DoT$ is a hybrid method where the joint learning is used along with an additional pruning loss.
$J\mbox{-}DoT$ achieves higher accuracy for similar efficiency, for both \wikisql and \wtq. For \tabfact the median is in the margin error but the best model using the joint learning outperforms the other learning strategies.}
\label{tab:results-loss}
\end{table*}

\section{All models results}
We report all $DoT$ results in Table~\ref{tab:all_results}. $C-DoT$ indicates the column based selection: For each token from one column, the pruning score attributes a column score instead of a token score. The column score is computed as an average of its tokens' scores.    
\begin{table*}
\centering
\resizebox{1.\textwidth}{!}{
\begin{tabular}{l*{3}{|lll}}
\textbf{Dataset} & \multicolumn{3}{|c}{\textbf{\wikisql}} & \multicolumn{3}{|c}{\textbf{\tabfact}} & \multicolumn{3}{|c}{\textbf{\wtq}} \\
\bottomrule
\textbf{Model}                                             & \textbf{test accuracy} & \textbf{Best}     & \textbf{NPE/s} &
                                                             \textbf{test accuracy} & \textbf{Best}     & \textbf{NPE/s} &
                                                             \textbf{test accuracy} & \textbf{Best}     & \textbf{NPE/s} \\
\bottomrule
\sota                                                      & \boldmath$83.9$        &       $-$            &          $-$      &
                                                             \boldmath$81.0$        &       $-$            &          $-$      &
                                                             \boldmath$51.8 \pm 0.6$& \boldmath$52.3$   & $-$  \\
\bottomrule                                                           
$\text{\ \ \ \ }CC\xrightarrow[]{1024} \mbox{TAPAS}(s)$    &$74.6 \pm 0.1$          & $74.69$           & $3900$         &
                                                            $73.3 \pm 0.1$          & $73.40$           & $4400$         &
                                                            $36.0 \pm 0.4$          & $36.92$           & $3800$\\
$\text{\ \ \ \ }CC\xrightarrow[]{1024} \mbox{TAPAS}(m)$    &$81.6 \pm 0.2$          & $81.55$           & $2050$         &
                                                            $75.1 \pm 0.2$          & $75.75$           & $2300$         &
                                                            $42.9 \pm 0.3$          & $43.67$           & $2020$\\
\bottomrule
$\text{\ \ \ \ } CC \text{ }\xrightarrow[]{256}\tapas(l)$&  $76.4 \pm 0.3$          &  $77.15$          & $1870$         &
                                                            $75.1 \pm 0.3$          &  $76.13$          & $1900$       &
                                                            $44.8 \pm 0.5$          &  $45.47$          & $1900$ \\
$\text{\ \ \ \ } CC \text{ }\xrightarrow[]{512} \tapas(l)$ & $83.6 \pm 0.1$          & $83.65$          & $800$          &
                                                             $81.3 \pm 0.2$          & $81.60$          & $870$          &
                                                             $52.2 \pm 0.5$          & $52.74$          & $810$ \\
$\text{\ \ \ \ } CC \xrightarrow[]{1024}\tapas(l)$         & \boldmath$86.0 \pm 0.3$ & \boldmath$86.6$  & \boldmath$270$ &
                                                             $81.6 \pm 0.1$          & $81.64$          & $300$          &
                                                             $53.9 \pm 0.2$          & $54.30$          & $270$ \\
$HEM \xrightarrow[]{256} \tapas(l)$        &                 $ 77.4 \pm 0.3 $        &  $77.97$         & $1870$  &
                                                             $ 75.5 \pm 0.2 $        &  $75.80$         & $1900$  &
                                                             $ 47.3 \pm 0.1 $        &  $47.70$         & $1900$  \\
$HEM \text{ }\xrightarrow[]{512} \tapas(l)$ &                $83.8 \pm 0.4$          & $84.75$          & $800$          &
                                                            \boldmath$82.0 \pm 0.3$  & \boldmath$82.07$ & \boldmath$870$ &
                                                            $52.7 \pm 0.4$           & $53.61$          & $810$ \\
$HEM \xrightarrow[]{1024} \tapas(l)$        &               \boldmath$85.9 \pm 0.0$  & \boldmath$85.94$ & \boldmath$270$ &
                                                            $80.6 \pm 0.0$           & $80.6$           & $300$          &
                                                            \boldmath$54.0 \pm 0.9$  & \boldmath$54.93$ & \boldmath$270$ \\
\bottomrule
$\text{\ \ \ \ } CC \xrightarrow[]{1024}DoT(mini \xrightarrow[]{256} l)$        & $72.8 \pm 0.8$            & $73.01$            &  $1600$           &
                                                                                  $77.2 \pm 0.5$            & $77.72$            &  $1670$           &
                                                                                  $37.4 \pm 0.9$           & $39.48$           & $1600$\\
$\text{\ \ \ \ } CC \xrightarrow[]{1024}DoT(s\text{ } \xrightarrow[]{256} l)$ &  $74.2 \pm 3.6$           & $84.27$           & $1250$ &
                                                                                  $81.0 \pm 0.1$           & $81.17$           & $1300$ &  $40.8 \pm 0.4$           & $42.15$          & $1250$ \\
$\text{\ \ \ \ } CC \xrightarrow[]{1024}DoT(m \xrightarrow[]{256} l)$         &  $83.6 \pm 0.5$           & $84.67$           & $950$             &
                                                                                  $79.0 \pm 0.5$           & $81.28$           & $930$             & $42.4 \pm 0.5$           & $43.44$           & $950$\\
                                                                                  
$HEM \xrightarrow[]{1024}DoT(mini\text{ } \xrightarrow[]{256} l)$            &    $73.4 \pm 0.1$            & $73.70$           & $1600$            &
                                                                                  $77.6 \pm 0.2$            & $78.19$           & $1670$            &
                                                                                  $39.2 \pm 0.3$            & $39.8$           & $1600$           \\
$HEM \xrightarrow[]{1024}DoT(s\text{ } \xrightarrow[]{256} l)$                &  \boldmath$85.3 \pm 0.4$  & \boldmath$85.76$  & \boldmath$1250$   &
                                                                                  \boldmath$81.6 \pm 0.3$  & \boldmath$81.74$  & \boldmath$1300$   &
                                                                                    $42.1 \pm 0.7$        &      $42.15$          & $1250$\\
$HEM \xrightarrow[]{1024}DoT(m \xrightarrow[]{256} l)$                        &  \boldmath$85.5 \pm 0.2$  & \boldmath$85.82$  & \boldmath$950$   &
                                                                                  \boldmath$81.8 \pm 0.0$  & \boldmath$81.94$   & \boldmath$930$   &
                                                                                 $48.2 \pm 1.8$     &   $48.46$         & $950$\\
\bottomrule
$\text{\ \ \ \ } CC \xrightarrow[]{1024}C-DoT(mini\text{ } \xrightarrow[]{256} l)$ & $72.0 \pm 1.1$            & $74.06$           & $1560$            &
                                                                                  $77.3 \pm 4.8$            & $77.61$           & $$1600$$            &
                                                                                  $ 41.0 \pm 0.4$            & $42.13$           & $1560$            \\
$\text{\ \ \ \ } CC \xrightarrow[]{1024}C-DoT(s\text{ } \xrightarrow[]{256} l)$ & $74.3 \pm 0.1$            & $74.49$           & $1250$            &
                                                                                  $77.2 \pm 4.9$            &$78.04$            &  $1300$           &
                                                                                  $48.1 \pm 2.4 $           & $49.47$           & $1250$\\
$\text{\ \ \ \ } CC \xrightarrow[]{1024}C-DoT(m \xrightarrow[]{256} l)$         & $73.9 \pm 0.2$            & $7454$            & $ 950$            &
                                                                                  $78.3 \pm 0.7 $           & $80.12$           & $ 930$            &
                                                                                  \boldmath$50.1\pm 0.5$  & \boldmath$ 50.14$   & \boldmath$950$\\
$HEM \xrightarrow[]{1024}C-DoT(mini\text{ } \xrightarrow[]{256} l)$            &  $72.0 \pm 1.1$            & $74.06$           & $1560$            &
                                                                                  $ 78.1 \pm 4.8$            & $78.13$           & $$1600$$            &
                                                                                  $ 41.5 \pm 0.1$            & $41.99$           & $1560$            \\
$HEM \xrightarrow[]{1024}C-DoT(s\text{ } \xrightarrow[]{256} l)$                & $74.5 \pm 0.8$            &  $74.72$          & $1250$            &
                                                                                      $58.5 \pm 0.3$                      &   $59.19$                &$1300$             &
                                                                                $40.9 \pm 0.2$        &      $41.23$          & $1250$\\
$HEM \xrightarrow[]{1024}C-DoT(m \xrightarrow[]{256} l)$                        & $74.3 \pm 2.1$            &  $74.56$          & $950$             &
                                                                                         $77.3 \pm 0.1$                   &     $77.71$              &$930$              &
                                                                                  $40.1 \pm 2.4$  &   $49.13$         & $950$\\
\end{tabular}
}
\caption{Summary of all the experiments' results on the accuracy efficiency trade-off. The \sota (detailed in Appendix~\ref{app:sota}) correspond to the values of \cite{min-etal-2019-discrete} for \wikisql, \cite{eisenschlos2020understanding} for \tabfact and \cite{yin2020tabert} for \wtq. For each dataset, the \sota, the best baseline model on accuracy, and the $DoT$ models that reach the best accuracy efficiency trade-off are highlighted.}
\label{tab:all_results}
\end{table*}

\end{document}